# Generalizing Adversarial Examples by AdaBelief Optimizer

Yixiang Wang, Jiqiang Liu, and Xiaolin Chang

*Abstract*—Recent research has proved that deep neural networks (DNNs) are vulnerable to adversarial examples—the legitimate input added with imperceptible and well-designed perturbations can fool DNNs easily in the testing stage. However, most of the existing adversarial attacks are difficult to fool adversarially trained models. To solve this issue, we propose an *AdaBelief iterative Fast Gradient Sign Method* (AB-FGSM) to generalize adversarial examples. By integrating AdaBelief optimization algorithm to I-FGSM, we believe that the generalization of adversarial examples will be improved, relying on the strong generalization of AdaBelief optimizer. To validate the effectiveness and transferability of adversarial examples generated by our proposed AB-FGSM, we conduct the white-box and black-box attacks on various single models and ensemble models. Compared with state-of-the-art attack methods, our proposed method can generate adversarial examples effectively in the white-box setting, and the transfer rate is 7%~21% higher than latest attack methods.

## I. INTRODUCTION

In the past few years, the rise of artificial intelligence flows from the advancement of deep learning (DL) technologies. DL has made significant success and state-of-the-art performance in various domains [1]–[3].

However, the emergency of adversarial examples [4] raises hindrance to DL techniques and their practical applications, and the security and reliability of DL techniques challenge their users. In particular, adversarial examples are the legitimate input coupled with well-designed and imperceptible perturbations that will fool deep neural networks (DNNs) and result in the less convincing results of DNNs. And the characteristic of adversarial examples, transferability, makes adversarial examples more aggressive and generalized. Transferability means that adversarial examples generated by one model can fool other models. Adversarial examples can be a two-edged sword. It can help recognize the models' vulnerability and improve their robustness by adversarial training before they are released. It is critical to learn how to generate high generalized adversarial examples and it has been studied in the literature [5]–[7].

There are also two types of attacks according to the adversarial knowledge: white-box attack and black-box attack. In the white-box settings, the adversary has detailed knowledge of the model it wants to attack, as well as the data information. As for black-box attacks, the adversary has no information of the targeted model, but it can draw support from the transferability of adversarial examples to attack the targeted model. With the transferability, the adversary can implement a black-box attack.

Previously, researchers proposed first-order gradient-based one-step [7] or multiple-step [8] adversarial attacks to generate adversarial examples in the white-box setting. However, they only paid attention to how to generate adversarial examples effectively. Adversarial examples generated by these attacks have a low success rate on the adversarially trained models or the ensemble trained models [9] or models with other defensive mechanisms such as input transformation [10][11] under the black-box setting. Especially, the transferability of adversarial examples proposed by recent work did not perform as well as they claimed, whose results will be shown in Section IV.

In this paper, we focus on the transferability of adversarial examples in the black-box setting. Concretely, we propose an iterative method to improve the transferability of adversarial examples in the black-box setting and maintain the success rate in the white-box setting: AdaBelief iterative Fast Gradient Sign Method [12].

Inspired by the fact that AdaBelief adaptive optimizer is currently the best optimizer from the aspect of convergence and generalization, we propose to integrate it to the I-FGSM[8] method to improve the transferability of adversarial examples. Extensive experimental results verify that our proposed method can generate adversarial examples with a stable and high white-box success rate and black-box success rate, compared with the latest methods.

The rest of the paper is organized as follows. In Section II, we present the notations, and describe the background and relative methods of adversarial examples. The detail description of AB-FGSM is provided in section III. Section IV analyzes the experimental results. Conclusion and future work are investigated in Section V.

## II. BACKGROUNDS

In this section, we give a detailed explanation of basic notations and Fast Gradient Sign Method (FGSM). On this

Yixiang Wang, Jiqiang Liu (Corresponding author) and Xiaolin Chang are with Beijing Key Laboratory of Security and Privacy in Intelligent Transportation, Beijing Jiaotong University, Beijing 100044, China (e-mail: {18112047, jqliu, xlchang}@bjtu.edu.cn).

basis, the attack families based on FGSM are extended, such as MI-FGSM, NI-FGSM, AI-FGSM. Given a Deep Neural Network (DNN) classifier $f(x;\theta): x \in \mathcal{X} \rightarrow y \in \mathcal{Y}$, one can output a prediction label $y$ corresponding to an input $x$, and the input has a ground truth label $C^*(x)$. In the adversarial settings, an adversary wants to find an adversarial example $x'$ that is almost the same $x$ but has imperceptible differences. According to the adversarial goals, there exist two types of adversarial examples: non-targeted and targeted. Specifically, the prediction label of non-targeted adversarial examples is $f(x') \neq C^*(x)$, and $f(x') = y' \neq C^*(x)$ is the prediction label of targeted adversarial examples. In this paper, we pay attention to the prior work on non-targeted adversarial examples.

*A. Fast Gradient Sign Method*

Szegedy *et al.*[4] were the first to exploit gradient information to generate adversarial examples, and the proposed white-box attack is called Fast Gradient Sign Method (FGSM). Its purpose is to find an adversarial example $x'$ such that the loss function value $J(x', C^*(x))$ is maximized. Here, the loss function is cross-entropy. It is worth noting that the difference between legitimate $x$ and adversarial $x'$ should be within the $L_\infty$ ball around $x$, i.e., $\|x - x'\|_\infty \leq \epsilon$. The formula is

$$x' = x + \epsilon \cdot sign(\nabla_x J(x, C^*(x)))$$

where $\nabla_x J(x, C^*(x))$ denotes the derivative of loss function $J(\cdot)$ w.r.t the input $x$. The limitation of FGSM is that it only uses one-step gradient information to generate adversarial examples, which leads to a low generation rate.

*B. Optimization Algorithm*

Neural networks are trained with first-order gradient descent algorithms. There exist two families of gradient descent algorithms: accelerated stochastic gradient descent (SGD) family, such as momentum[13] and Nesterov[14]; adaptive learning rate family[12], such as Adam[15], Adadelta[16]. The two families have their own advantages and disadvantages. Concretely, DNNs trained with SGD family has a strong generalization ability, but the convergence rate is low. The adaptive family is the opposite of SGD family. They train DNNs faster with the cost of generalization ability. Researchers who want to improve the transferability of adversarial examples try to seek a solution in these optimization families.

*C. Iterative-FGSM and Its Family*

To solve the low generation rate of FGSM, Iterative-FGSM (**I-FGSM**) was proposed in [8]. Compared with FGSM, I-FGSM is a multi-step adversarial attack. The iterative function is expressed as:

$$x'_0 = x, \; x'_{t+1} = clip_x^\epsilon \left( x'_t + \alpha \cdot sign(\nabla_x J(x, C^*(x))) \right)$$

This means the adversary allocates the total perturbations into $T$ iterations, i.e., every $\alpha = \epsilon/T$ perturbations are added to the input in one iteration. $clip_x^\epsilon(\cdot)$ function restricts the input into the $\epsilon$- ball around $x$. Experimental results showed that the multi-step attack is more aggressive and stronger than the one-step attack.

Since then, researchers have devoted themselves to investigating how to generate adversarial examples with more aggression and transferability on I-FGSM, especially from optimization algorithms. Dong *et al.*[5] integrated momentum accelerated gradient [13] into the I-FGSM to stabilize the iterative directions, and the proposed attack, **MI-FGSM**, has a stronger transferability for adversarial examples. The procedure is formalized as:

$$g_0 = 0, \; g_{t+1} = \mu \cdot g_t + \frac{\nabla_x J(x, C^*(x))}{\|\nabla_x J(x, C^*(x))\|_1}$$

$$x'_{t+1} = clip_x^\epsilon \left( x'_t + \alpha \cdot sign(g_{t+1}) \right)$$

where $g_t$ in is the accumulated gradients of the first $t$ iterations with a decay factor $\mu$. Then, $g_t$ is adopted to the I-FGSM.

After that, Lin *et al.*[6] adapted the Nesterov accelerated gradient [14] to the I-FGSM since it is better than momentum for conventional optimization, which we call **NI-FGSM**. The update formulas that differ from momentum is shown below:

$$x_t^{nes} = x'_t + \alpha \cdot \mu \cdot g_t$$

$$g_{t+1} = \mu \cdot g_t + \frac{\nabla_x J(x_t^{nes}, C^*(x))}{\|\nabla_x J(x_t^{nes}, C^*(x))\|_1}$$

Previous work only focused on the view of SGD families. Yin *et al.*[17] combined adaptive Adam method [15] with I-FGSM, **AI-FGSM**. The formula is shown below:

$$g_t = \frac{\nabla_x J(x_t^{nes}, C^*(x))}{\|\nabla_x J(x_t^{nes}, C^*(x))\|_1}$$

$$m_{t+1} = \beta_1 \cdot m_t + (1 - \beta_1) \cdot g_t, \; v_{t+1} = \beta_2 \cdot v_t + (1 - \beta_2) \cdot g_t^2$$

$$\tau_{t+1} = \frac{m_{t+1}}{\delta + \sqrt{v_{t+1}}}$$

$$\alpha_{t+1} = \alpha \cdot \frac{\sqrt{1 - \beta_2^{i+1}}}{1 - \beta_1^{i+1}} \bigg/ \sum_{i=0}^{T-1} \frac{\sqrt{1 - \beta_2^{i+1}}}{1 - \beta_1^{i+1}} \quad \text{(II.1)}$$

$$x'_{t+1} = x'_t + \alpha_{t+1} \cdot \frac{\tau_{t+1}}{\|\tau_{t+1}\|_2}$$

where $m_t$ and $v_t$ denote the exponential moving average (EMA) of $g_t$ and $g_t^2$. $\beta_1$ and $\beta_2$ is exponential decay rates. The perturbation is denoted as $\tau_t$. This is a variant of standard adaptive Adam method. The difference focuses on the Eq.(II.1)(II.1). The standard Adam method does not have the term $\sum_{i=0}^{T-1} \frac{\sqrt{1-\beta_2^{i+1}}}{1-\beta_1^{i+1}}$, the authors added it since it can normalize the step size $\alpha$ in iterations. However, the authors did not use the $sign(\cdot)$ function to generate adversarial examples, which we think this is not reasonable and we correct this in the experiments.

All previous works have limitations. Although momentum and Nesterov make DNN models more generalized, the impact becomes very limited in the adversarial settings. And the adversarial examples generated by AI-FGSM do not generalize well since DNNs trained with Adam generalize poorly, which is in our expectation. To solve these limitations, we propose our AB-IFGSM.

### III. METHODOLOGY

In this paper, we propose a progressive AdaBelief iterative method for generating adversaria samples, which can attack both white-box and black-box models. In this section, we first describe how to combine AdaBelief optimizer with iterative Fast Gradient Sign Method. Then, we illustrate how our proposed method attacks single and ensemble models.

#### A. AdaBelief Iterative Fast Gradient Sign Method

AdaBelief optimizer is proposed to solve the problem that adaptive optimization methods cannot generalize as well as stochastic gradient descent(SGD) optimization methods [12]. AdaBelief optimizer solved this by adding a 'belief' parameter $1/\sqrt{s_t}$, easily modified from Adam optimizer. The detailed algorithms of Adam and Adabelief optimizer are shown below:

$$g_t = \nabla_\theta f(\theta), \quad m_t = \beta_1 m_{t-1} + (1-\beta_1)g_t, \quad \widehat{m_t} = \frac{m_t}{1-\beta_1^t}$$

$$v_t = \beta_2 v_{t-1} + (1-\beta_2)g_t^2, \quad \widehat{v_t} = \frac{v_t}{1-\beta_2^t} \quad \text{(III.1)}$$

Adam: $\quad \Delta\theta_t = \alpha \cdot \frac{1}{\sqrt{\widehat{v_t}}+\epsilon} \cdot \widehat{m_t}$ (III.2)

$$s_t = \beta_2 s_{t-1} + (1-\beta_2)(g_t - m_t)^2, \quad \widehat{s_t} = \frac{s_t + \varepsilon}{1-\beta_2^t} \quad \text{(III.3)}$$

AdaBelief: $\quad \Delta\theta_t = \alpha \cdot \frac{1}{\sqrt{\widehat{s_t}}+\epsilon} \cdot \widehat{m_t}$ (III.4)

The improvement of Adam method in the Adabelief method is shown in Eq.(III.3)(III.3) and Eq.(III.4)(III.4). The term $(g_t - m_t)^2$ promises that AdaBelief method can update parameters steadily on any surface. Especially when the surface is steep, the gradient $g_t$ is large, we want the parameters to be updated faster. In this case, $m_t$ and $v_t$ are large according to EMA theory. This will cause $1/\sqrt{v_t}$ in Eq.(III.2)(III.2) very small and the update will be slow. This phenomenon is the shortcoming of Adam method. However, in AdaBelief settings, $1/\sqrt{s_t}$ is large since $g_t$ and $m_t$ are close, which will lead to a fast convergence and agree with expectation. This is what we call 'belief', since $1/\sqrt{s_t}$ can reflect the belief of the gradient prediction: if $1/\sqrt{s_t}$ is small, then AdaBelief method believes in taking a small step to update parameters; otherwise AdaBelief takes a big step update.

**Algo. 1. Ab-FGSM**

1: **Input:** The classifier learnt function $f(\cdot)$ with the cross-entropy objective function $J(\cdot)$; a legitimate input $x$ and its ground-truth label $C^*(x)$; the total steps $T$ with each step $t$; the maximum perturbation $\epsilon$; AdaBelief factors includes exponential decay rates $\beta_1, \beta_2$, a denominator stability parameter $\varepsilon$.

2: **Output:** An adversarial example $x'$ with $\|x - x'\|_\infty < \epsilon$;

3: **Initialize** $m_0 \leftarrow 0$, $s_0 \leftarrow 0$, $t \leftarrow 0$, $x'_0 = x$

4: while $t < T$ do:

5:     $t \leftarrow t+1$

6:     $g_t \leftarrow \nabla_x J(\theta, x'_{t-1}, C^*(x))$

7:     $\gamma = \sum_{i=1}^{t} \frac{\sqrt{1-\beta_2^{i+1}}}{1-\beta_1^{i+1}}$

8:     $m_t \leftarrow \beta_1 m_{t-1} + (1-\beta_1)g_t$

9:     $s_t \leftarrow \beta_2 s_{t-1} + (1-\beta_2)(g_t - m_t)^2$

10:    **if** amsgrad:

11:       $s_t = \max(s_{t-1}, s_t)$

12:    $\widehat{m_t} \leftarrow \frac{m_t}{1-\beta_1^t}, \quad \widehat{s_t} \leftarrow \frac{s_t + \varepsilon}{1-\beta_2^t}$

13:    $x'_t = x'_{t-1} + \frac{\alpha}{\gamma} \cdot sign\left(\frac{\widehat{m_t}}{\sqrt{\widehat{s_t}}+\varepsilon}\right)$

14:    $x'_t = Clip_x^\epsilon(x'_t)$

15: **end while**

16: **return** $x'$

Intuitively, we integrate AdaBelief method to I-FGSM to improve adversarial examples' transferability because of the faster convergence rate and stronger generalization performance of AdaBelief method, and we call it AB-FGSM algorithm. The procedure of algorithm is shown in **Algo. 1**.

Specifically, AB-FGSM takes $T$ iterations to generate adversarial examples. In each iteration, we first compute the gradient of current input $x'$, as described in line 6. Then in lines 8, 9 and 12, we calculate the EMA of $g_t$ and $(g_t - m_t)^2$ the bias-corrected EMA, where there is no difference with the standard AdaBelief method.

The difference between AdaBelief and AB-IFGSM is in lines 7, 11, 13 adn 14. In line 7, inspired by the previous work of AI-FGSM[17], we also add a normalized term $\gamma$ to further set the step size. And our normalized term differs from theirs. We only add the $1 \sim t$-th normalized values, but theirs added all normalized values initially, which we think is counterintuitive. Our benefit is that only the previous normalized terms affect the step size, and the normalized terms that do not iterate have no effect on the current step size. We think this is more reasonable than theirs. In line 11, AMSGrad [18] skill is used in our algorithm to help AB-FGSM prevent converging to the sub-optimal point. Line 13, 14 are the description of FGSM.

We follow their experimental pattern [5], [6], [17] to validate two parts' efficiency and effectiveness: attacking one model and attacking ensemble models. In the next section, we will give a detailed description.

*B. Attacking Single and Ensemble Models*

As mentioned in Section I, in adversarial settings, there are two attack types: white-box attack and black-box attack. To validate the effectiveness and transferability of adversarial examples generated by our proposed AB-FGSM, we conduct the white-box and black-box attacks simultaneously.

Traditionally, white-box and black-box attacks are conducted in a single DNN model. We generate adversarial examples on a DNN model by our proposed AB-FGSM and transfer these examples to attack some DNN models we do not know. Dong *et al.*[5] emphasized that adversarial examples that can threaten an ensemble model have a stronger transferability. An ensemble model means the logit outputs of multiple DNN models are fused together with different weights. For instance, the ensemble model of $K$ models is shown:

$$L(x) = \sum_{k=1}^{K} w_k \cdot l_k(x) \quad (III.5)$$

where $w_i$ and $l_i(x)$ donate the weight and the logit outputs of $i$-th model. So, in the experiments, we not only attack a single DNN model in white-box and black-box manners, but the attack an ensemble model in two manners.

TABLE I. THE SUCCESS RATES (%) OF FIVE ADVERSARIAL ATTACK METHODS AGAINST SIX SINGLE MODELS ON WHITE-BOX AND BLACK-BOX SETTINGS

|  | Method | Inc-v3 | Inc-v4 | IncRes-v2 | Res-101 | Inc-v3$_{ens3}$ | Inc-v3$_{adv}$ |
|---|---|---|---|---|---|---|---|
| Inc-v3 | I-FGSM | **100.0** | 24.7 | 17.5 | 21.4 | 26.1 | 19.5 |
|  | MI-FGSM | **100.0** | 51.4 | 46.0 | 44.1 | **35.3** | 33.8 |
|  | NI-FGSM | **100.0** | 53.7 | 43.7 | 42.3 | 34.4 | 30.5 |
|  | AI-FGSM | **100.0** | 44.7 | 38.5 | 37.4 | 33.2 | 28.8 |
|  | AB-FGSM | **100.0** | **55.2** | **51.2** | **47.3** | *34.6* | **39.0** |
| Inc-v4 | I-FGSM | 38.5 | **99.8** | 18.8 | 24.2 | 28.5 | 21.4 |
|  | MI-FGSM | 65.7 | **99.8** | 51.4 | 49.9 | 37.5 | 36.4 |
|  | NI-FGSM | 70.1 | **99.8** | 54.4 | 51.8 | **38.7** | 36.8 |
|  | AI-FGSM | 71.8 | *99.7* | **56.9** | 54.6 | 37.1 | 43.9 |
|  | AB-FGSM | **72.2** | *99.7* | *56.0* | 53.2 | *37.8* | **44.3** |
| IncRes-v2 | I-FGSM | 40.0 | 35.8 | *98.2* | 29.7 | 30.8 | 24.1 |
|  | MI-FGSM | 69.3 | 61.5 | *98.7* | 52.5 | 39.1 | 40.1 |
|  | NI-FGSM | 71.9 | 65.5 | *99.0* | 55.6 | **41.6** | 40.6 |
|  | AI-FGSM | 64.4 | 58.1 | *98.3* | 49.7 | 38.6 | 38.2 |
|  | AB-FGSM | **76.6** | **68.5** | **100.0** | **58.6** | *41.0* | **45.0** |
| Res-101 | I-FGSM | 36.7 | 29.5 | 23.5 | *99.5* | 29.0 | 22.7 |
|  | MI-FGSM | 61.4 | 55.3 | 46.2 | *99.5* | **40.9** | 35.7 |
|  | NI-FGSM | 62.2 | 55.8 | 47.0 | *99.6* | 38.9 | 34.9 |
|  | AI-FGSM | 58.2 | 50.7 | 42.8 | *99.5* | 38.7 | 32.7 |
|  | AB-FGSM | **69.3** | **62.8** | **55.7** | *99.5* | 36.2 | **41.6** |

## IV. EXPERIMENTS AND ANALYSIS

In this section, extensive experiments are conducted to validate the efficiency and effectiveness of our proposed method. We first introduce the dataset, models, and hyperparameters used in the experiments in Section IV.A. Then, the result analysis of attacking single model and ensemble models is given in Section IV.B and IV.C.

*A. Setup*

We choose the NeurIPS 2017 non-target adversarial competition dataset (https://www.kaggle.com/c/nips-2017-non-targeted-adversarial-attack), which contains 1000 samples. Each sample has a unique label, and its size is $299 \times 299 \times 3$. This dataset can be seen as a tiny version of ImageNet. We choose the competition dataset instead of ImageNet because the competition dataset can avoid the randomness of picking data.

In the experiments, we consider six models containing normal trained models and adversarially-trained models. For normally trained models, we study four models: Inception v3 (Inc-v3) [19], Inception v4 (Inc-v4), Inception Resnet v2 (IncRes-v2) [20], Resnet v2-101 (Res-101) [21]. For adversarially trained models, we consider Inc-v3$_{ens3}$ model and Inc-v3$_{adv}$ model.

As for the comparison, we compare our method with four iterative FGSM variants: I-FGSM, MI-FGSM, NI-FGSM, and AI-FGSM, which are described in Section II. They are from the FGSM family and are the combinations of optimization methods and I-FGSM algorithm. We do not take FGSM into account because the previous work demonstrated its weakness in improving the transferability of adversarial examples.

In terms of hyper-parameters, we follow the part of the settings described in [5]. That is, the maximum perturbation $\epsilon$ is 16, the momentum factor $\mu$ is 1.0. The number of iterations $T$ in our experiments is set to 10. $\alpha$ in both MI-FGSM and NI-

FGSM is $\epsilon/T$. In AI-FGSM and AB-FGSM, exponential decay rates $\beta_1$ and $\beta_2$ are set to 0.99 and 0.999. $\varepsilon$ is a small number and is set to $10^{-14}$. The other hyper-parameters are the same as above.

In the last, we introduce the evaluation metric, success rate. The formula is shown below:

$$SuccRate = \frac{\# \text{ of adv samples}}{\# \text{ of all samples}}$$

It is reasonable to choose one metric to evaluate our experiment since we pay attention to the transferability of adversarial examples and do not care about the perturbations. From this point of view, the success rate is undoubtedly an intuitionistic and proper metric. Meanwhile, we call the white-box attack success rate as generation rate and the black-box attack success rate as transfer rate.

We implement our method in TensorFlow Python library. We use an Intel Xeon Silver 4114 CPU with a single NVIDIA TITAN Xp GPU for experiments.

*B. Analysis of Attacking Single Model*

TABLE I shows the results of attacking six single models by our proposed AB-FGSM, where underlined numbers denote the white-box results and normal numbers are the results of black-box attacks.

From the results, we can see that in the Inc-v3 model, all methods' generation rate is 100%, but the transfer rates are diverse from each other. Specifically, I-FGSM has the lowest transfer rate among the five methods. This meets the expectations since the vanilla optimization method only finds a point among the $\epsilon$-ball space around $x$ that can fool the model. The space contains the points that can fool the model and has a strong transferability. The other sophisticated optimization methods not only consider the generation rate but also ensure transferability. Among them, to our surprise, NI-FGSM performs worse than MI-FGSM, where the transfer rates of NI-FGSM on the IncRes-v2~ Inc-v3$_{adv}$ are 1%~4% lower than that of MI-FGSM. However, our AB-FGSM method performs best in terms of both generation rate and transfer rate. Basically, the transfer rate is almost 10%~38% larger than that of I-FGSM and 3%~6% larger than that of the second-best MI-FGSM.

As for the Inc-v4 model, the results are different from that of Inc-v3. The same conclusion is that adversarial examples generated by I-FGSM have a limited transferability. And the generation rates of all methods are almost the same. However, five methods' transfer rates are increased gradually from I-FGSM to AB-FGSM except in the IncRes-v2 and Inc-v3$_{ens3}$ models. In these two models, our method achieves the second-best scores. NI-FGSM performs better than MI-FGSM, which implies that NI-FGSM performs well in complicated models.

Another interesting thing is that the performance of AI-FGSM is slightly worse than that of ours, which infers that AI-FGSM is not as stable as our AB-FGSM method. Remarkably, our AB-FGSM gets an acceptable performance. Our methods achieve the best transferability on Inc-v3, Res101, and Inc-v3$_{adv}$ models. And compared with the second-best AI-FGSM, the improvement is only about 1%, which is limited. But ours is still better than MI-FGSM and NI-FGSM.

The performance of five methods on IncRes-v2 and Res-101 models is relatively consistent. Notably, NI-FGSM outperforms MI-FGSM, further validating our implication: NI-FGSM performs better than MI-FGSM as the model becomes complex. AI-FGSM performs poorly on these two models, even worse than MI-FGSM. This further illustrates the unreliability of AI-FGSM. Our proposed method achieves 100% generation rate and the strongest transferability, whose scores are 3%~5% larger than that of the second-best NI-FGSM method. As for the Res-101 model, the generation rates of five methods are almost the same. Our method gets the best transferability scores among four models in the black-box setting, which are about 7% larger than that of the second-best NI-FGSM.

To sum up, when attacking a single model, our proposed method, AB-FGSM, can maintain a high generation rate and the adversarial examples generated by it have strong transferability. We also find one intriguing phenomenon from the experimental results: adversarial examples generated by AI-FGSM are not transferable all the time. In the next sub-section, we will analyze the performance of five methods when attacking ensemble models.

TABLE II. THE SUCCESS RATES (%) OF FIVE ADVERSARIAL ATTACK METHODS AGAINST ENSEMBLE MODELS ON WHITE-BOX AND BLACK-BOX SETTINGS

|  | Method | -Inc-v3 | -Inc-v4 | -IncRes-v2 | -Res-101 | -Inc-v3$_{ens3}$ | -Inc-v3$_{adv}$ | Avg. |
|---|---|---|---|---|---|---|---|---|
| Ensemble | I-FGSM | 96.7 | 97.0 | **98.7** | 95.8 | 96.5 | 98.2 | 97.2 |
|  | MI-FGSM | 96.3 | 96.5 | 98.6 | 96.2 | 96.3 | 97.3 | 96.9 |
|  | NI-FGSM | 97.9 | **98.2** | 98.0 | **98.1** | **99.5** | 98.7 | 98.4 |
|  | AI-FGSM | 96.2 | 96.7 | 98.4 | 95.6 | 95.2 | 97.9 | 96.7 |
|  | AB-FGSM | **98.6** | **98.2** | 98.5 | *98.0* | *99.4* | **99.1** | **98.6** |
| Hold-out | I-FGSM | 55.0 | 45.8 | 39.4 | 36.1 | 25.6 | 24.8 | 37.8 |
|  | MI-FGSM | 76.3 | 73.6 | 68.8 | 64.0 | 34.5 | 37.3 | 59.1 |
|  | NI-FGSM | 52.7 | 50.6 | 56.1 | 44.6 | 32.9 | 30.4 | 44.6 |
|  | AI-FGSM | 67.2 | 63.6 | 58.5 | 53.2 | 29.9 | 37.9 | 51.7 |
|  | AB-FGSM | **85.2** | **82.2** | **85.2** | **69.7** | **36.5** | **38.7** | **66.3** |

*C. Analysis of Attacking Ensemble Model*

In this section, we analyze the performance of five methods attacking ensemble models. We give an example to demonstrate the TABLE II. In the column '-Inc-v3', 'Ensemble' denotes that multiple models containing the other five models except the Inc-v3 model are fused together by the Eq.(III.5)(III.5). 'Hold-out' denotes that adversarial examples generated by one method on the ensemble model are transferred to the Inc-v3 model.

As shown in TABLE II, in the ensemble white-box attack setting, every method works well. Concretely, our method achieves the best generation rate on the three ensemble models and the second-best in two ensemble models, where two second-best generation rates are only 0.1% lower than the best. What is surprising is that I-FGSM achieves the best generation rate on the '-IncRes-v2' ensemble model. However, this is acceptable because there exists the difference of generation rates between I-FGSM and ours. NI-FGSM achieves the best generation rates on three ensemble models, but one of them is the same as ours. The others are 0.1% larger than ours. Other methods, such as MI-FGSM and AI-FGSM, are slightly inferior to ours.

The transferability rates of the five methods on ensemble models are widely divergent and NI-FGSM is beyond our expectation. Specifically, NI-FGSM does not perform much better than I-FGSM. The performances of AI-FGSM and MI-FGSM are superior to NI-FGSM. It is an intriguing phenomenon since NI-FGSM performs well in attacking a single model, whatever white-box and black-box. Our proposed AB-FGSM still performs the best in the five methods, where the transfer rate of our generated algorithm is the best in any ensemble model.

In conclusion, our proposed AB-FGSM can generate adversarial examples steadily, whether in the single model or in the ensemble model. Meanwhile, the adversarial examples generated by ours can transfer to other models with a high attack success rate.

## V. CONCLUSIONS

In this paper, we propose an iterative method for generating adversarial samples, AB-FGSM, which can be used to conduct white-box attacks to DNNs. The extensive experimental results verify that our algorithm can generate adversarial examples with a higher generation rate in the white-box setting and higher transferability rates in the black-box setting. The transfer rate is 7%~21% higher than that of the latest attack methods.